\documentclass{article}
\pdfoutput=1

\usepackage{arxiv}

\usepackage[utf8]{inputenc} 
\usepackage[T1]{fontenc}    
\usepackage{hyperref}       
\usepackage{url}            
\usepackage{booktabs}       
\usepackage{amsfonts}       
\usepackage{nicefrac}       
\usepackage{microtype}      
\usepackage{lipsum}		
\usepackage{graphicx}
\usepackage{amsmath}
\usepackage{natbib}
\usepackage{doi}
\usepackage{makecell}
\usepackage{color}
\usepackage{xcolor}

\title{Technical Report of Team GraphMIRAcles in the WikiKG90M-LSC Track of OGB-LSC @ KDD Cup 2021}

\author{Jianyu Cai \qquad Jiajun Chen \qquad Taoxing Pan \qquad Zhanqiu Zhang \qquad Jie Wang\thanks{\ \ Corresponding author.} \\
  University of Science and Technology of China \\
  \texttt{\{jycai,jiajun98,tx1997,zqzhang\}@mail.ustc.edu.cn} \\
  \texttt{jiewangx@ustc.edu.cn} 
}

\renewcommand{\shorttitle}{Technical Report of Team GraphMIRAcles}

\hypersetup{
pdftitle={Technical Report for the WikiKG90M-LSC track of the OGB-LSC},
pdfsubject={q-bio.NC, q-bio.QM},
pdfauthor={David S.~Hippocampus, Elias D.~Striatum},
pdfkeywords={First keyword, Second keyword, More},
}

\begin{document}
\maketitle

\begin{abstract}
Link prediction in large-scale knowledge graphs has gained increasing attention recently. The OGB-LSC team presented OGB Large-Scale Challenge (OGB-LSC), a collection of three real-world datasets for advancing the state-of-the-art in large-scale graph machine learning. In this paper, we introduce the solution of our team GraphMIRAcles in the WikiKG90M-LSC track of OGB-LSC @ KDD Cup 2021. In the WikiKG90M-LSC track, the goal is to automatically predict missing links in WikiKG90M, a large scale knowledge graph extracted from Wikidata. To address this challenge, we propose a framework that integrates three components---a basic model ComplEx-CMRC, a rule miner AMIE 3, and an inference model to predict missing links. Experiments demonstrate that our solution achieves an MRR of 0.9707 on the test dataset. Moreover, as the knowledge distillation in the inference model uses test tail candidates---which are unavailable in practice---we conduct ablation studies on knowledge distillation. Experiments demonstrate that our model without knowledge distillation achieves an MRR of 0.9533 on the full validation dataset. 

\end{abstract}

\section{Introduction}
Knowledge Graphs (KGs) incorporate world knowledge with nodes and edges being entities and relations among them, respectively. Although knowledge graphs have made great achievements in many areas, they often suffer from the incompleteness problem, i.e., a lot of links between entities are missing. Therefore, link prediction--which aims to predict missing links in knowledge graphs---has drawn much attention in recent years.

Recently, many researchers focus on predicting missing links by only using the graph structure information \citep{complex, rotate, hake, dura}. Some other researchers incorporate text information to assist link prediction \citep{10.5555/3060621.3060801, kgbert, kim-etal-2020-multi}. However, these methods often have poor scalability to large knowledge graphs as they jointly learn text and graph representations, which are time-consuming. Therefore, it is still challenging to effectively predict missing links based on a pretrained text encoder (such as BERT and RoBERTa). Besides, it is also desirable to incorporate traditional rule mining methods, ensemble methods and knowledge distillation methods into link prediction, which are rarely discussed but largely benefit the model performance.

To address the above challenges, we propose a framework that contains three components---a basic model ComplEx-CMRC, a rule miner AMIE 3, and an inference model that integrates ensemble and knowledge distillation methods. To show the effectiveness of our model, we conduct experiments on the WikiKG90M-LSC dataset in the 2021 KDD Cup on OGB Large-Scale Challenge \citep{hu2021ogblsc}. Experiments demonstrate that our solution achieves an MRR of 0.9707 on the test dataset.

\section{Related Work}
\textbf{Knowledge Graph Embedding} \hspace{1.5mm} Knowledge graph embedding (KGE) has been shown to be a promising direction for link prediction \citep{complex, hake, dura, tact}. The key idea of KGE is to embed entities and relations of a KG into continuous vector space while preserving the graph structure information \citep{kge-survey}. Many KGE models, such as ComplEx \citep{complex}, formulate the link prediction task as a tensor completion problem. ComplEx introduces complex-valued embeddings for KGE, which can handle a large variety of binary relations, including symmetric and antisymmetric relations \citep{complex}. Some works incorporate text information into KGE models to improve the performance of link prediction \citep{kgbert, kim-etal-2020-multi}.

\textbf{Rule Mining} \hspace{1.5mm} Rule mining aims at learning logical rules based on observed co-occurrence patterns of relations \citep{amie+}. AIME 3 \citep{amie3} is an effective method for mining rules from a large-scale knowledge graph, which employs a number of sophisticated pruning strategies and optimizations.

\textbf{Ensemble} \hspace{1.5mm} Ensemble \citep{ensemble-tpami} is one of the most powerful techniques in practice to improve the performance of deep learning models. By simply averaging the output of a few independently trained neural networks over the same training data set, it can significantly boost the prediction accuracy over the test set comparing to each individual model \citep{allen-zhu-analysis}.

\textbf{Knowledge Distillation} \hspace{1.5mm} Knowledge distillation \citep{distillation} is a technique to transfer the knowledge from the cumbersome model to a small model that is more suitable for deployment. The superior performance of the ensemble model can also be distilled into a single model using knowledge distillation \citep{allen-zhu-analysis}.

\section{Method}
In this part, we introduce our proposed method in detail. In Section \ref{sec:overall_architecture}, we introduce the overall architecture of our method. In section \ref{sec:complex_cmrc}, \ref{sec:rule_miner}, and \ref{sec:inference_model}, we introduce the three components of our method, respectively.

\begin{figure*}[h]
  \centering
  \includegraphics[scale=0.35]{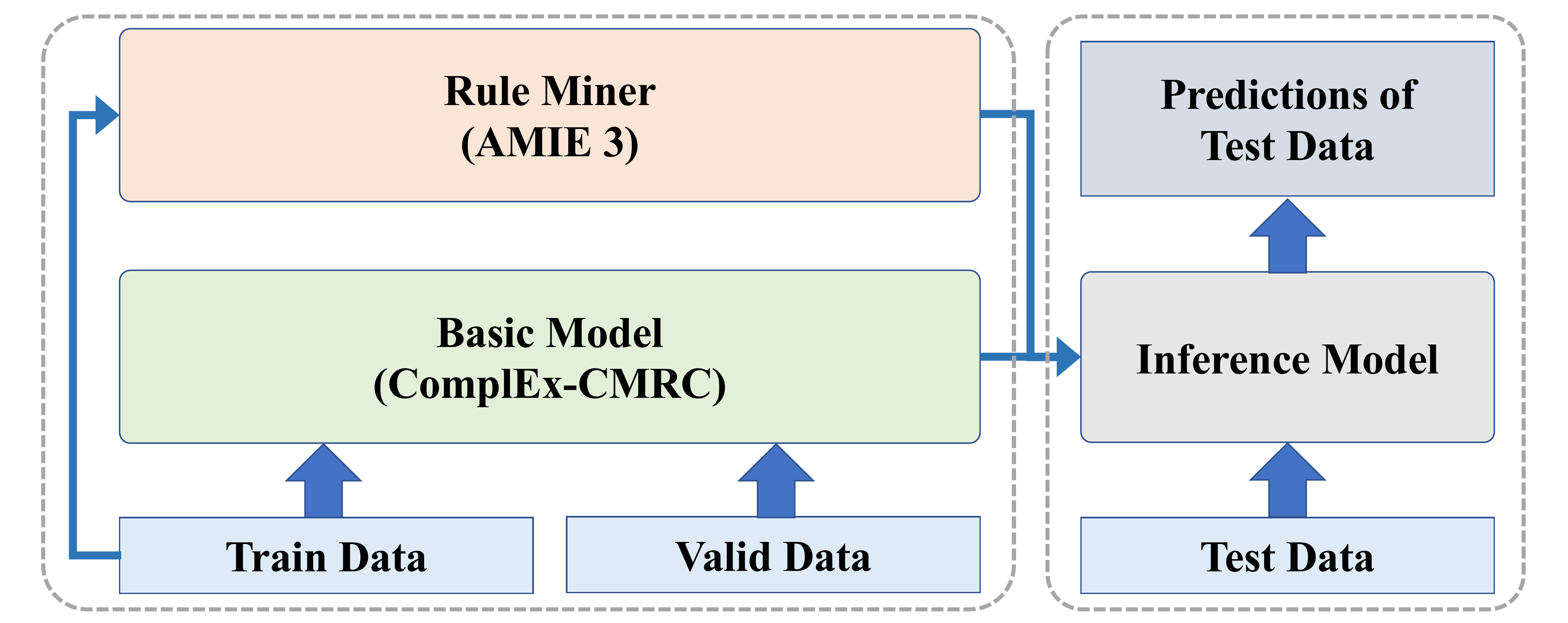}
  \caption{Overall architecture of our method.}
  \label{fig:overall_arch}
\end{figure*}

\subsection{Overall Architecture}\label{sec:overall_architecture}
The overall architecture of our method is shown in Figure \ref{fig:overall_arch}. Our method contains three components---the basic model ComplEx-CMRC, the rule miner AMIE 3 and the inference model. First, we train the basic models ComplEx-CMRC on the training dataset, and select the best ones based on their performances on the valid dataset. Then, we apply the rule miner AMIE 3 to generate Horn rules based on the training dataset. Finally, based on the basic models and the generated rules, we build an inference model to make predictions on the given test data.

\subsection{The ComplEx-CMRC Model}\label{sec:complex_cmrc}
To fully exploit the semantic information embedded in RoBERTa features and the structural information embedded in shallow features, we propose a model named ComplEx-CMRC, in which \textbf{CMRC} is our proposed encoder and \textbf{ComplEx} is the decoder. ``CMRC'' is the abbreviation for \textit{Concat-MLP with Residual Connection}.

\begin{figure*}[h]
  \centering
  \includegraphics[scale=0.35]{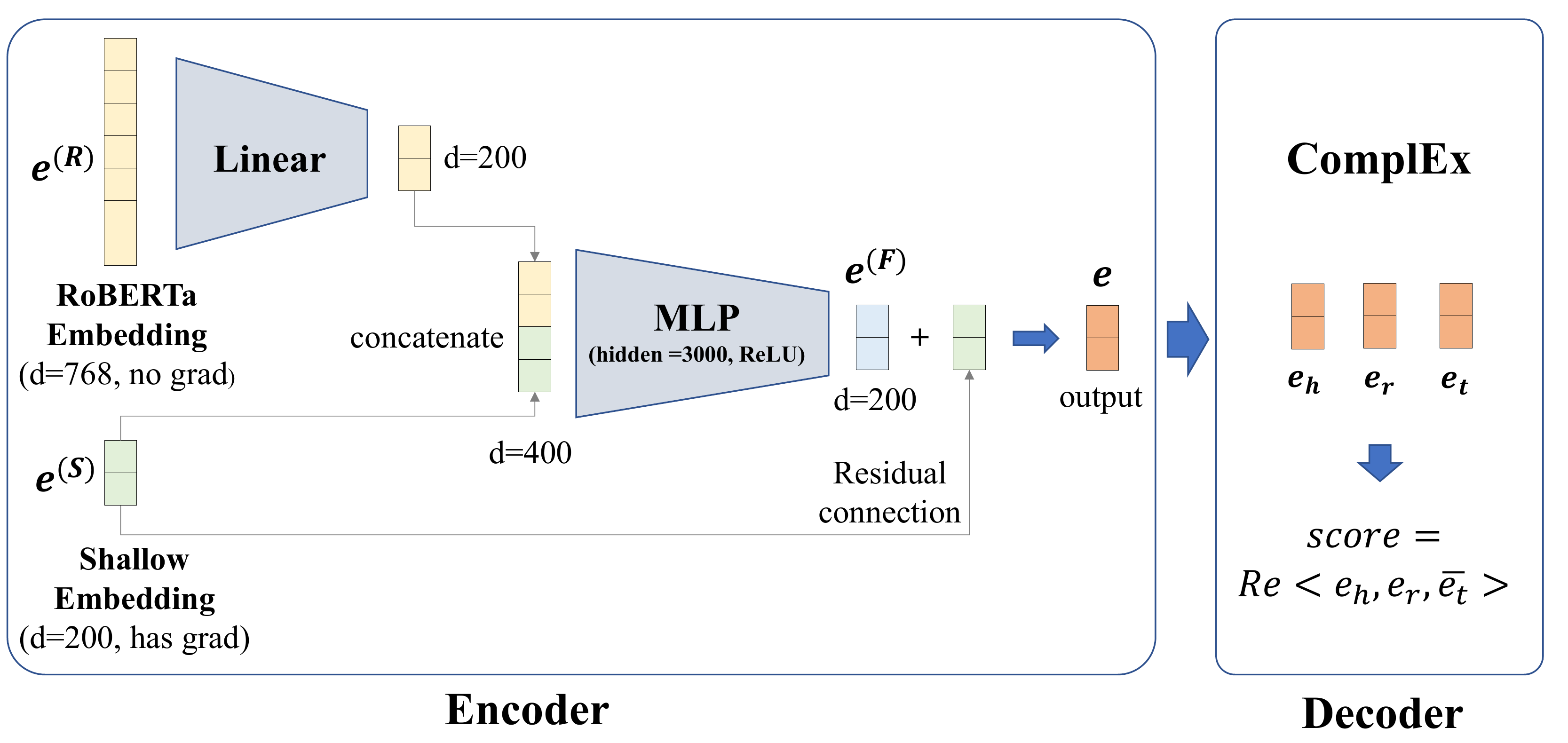}
  \caption{Overview of the ComplEx-CMRC model. ComplEx-CMRC consists of two parts---a CMRC encoder and a ComplEx decoder.}
  \label{fig:model_overview}
\end{figure*}

\subsubsection{Encoder: Concat-MLP with Residual Connection (CMRC)}
In this part, we introduce the proposed encoder CMRC. We use the same encoder architecture but two sets of parameters for entities and relations, respectively. For an arbitrary entity or relation, we are given the RoBERTa embedding $\mathbf{e}^{(R)} \in \mathbb{R}^{768}$, which encodes the semantic information. To capture the structural information in the KG, we define a shallow embedding $\mathbf{e}^{(S)} \in \mathbb{R}^{d}$ for each entity and relation, where $d < 768$ is the dimension size.

First, we use a linear layer to project $\mathbf{e}^{(R)} \in \mathbb{R}^{768}$ to $\mathbb{R}^{d}$. Then, we fuse the two kinds of embeddings by first concatenating them and then encoding the concatenated embedding with an MLP. That is, the fused embedding $\mathbf{e}^{(S)} \in \mathbb{R}^{d}$ is obtained by
\begin{align}
    \mathbf{e}^{(F)} = \text{MLP}([\text{Linear}(\mathbf{e}^{(R)}), \mathbf{e}^{(S)}]),
\end{align}
where $[\cdot, \cdot]$ denotes embedding concatenation and $\text{MLP}(\cdot)$ denotes a multi-layer perceptron with one hidden layer.

Finally, we apply residual connection to enable direct gradient flow to the shallow embeddings. That is, the final output of the encoder $e$ is obtained by
\begin{align}
    \mathbf{e} = \mathbf{e}^{(F)} + \alpha \mathbf{e}^{(S)},
\end{align}
where $\alpha \in \mathbb{R}$ is a trainable weight parameter.

\begin{table}[ht]
    \caption{The design choices of encoders. In the table, we list four types of encoders.}
    \label{table:encoder_type}
    \centering
    \resizebox{0.7 \columnwidth}{!}{
    \begin{tabular}{l *{1}{c}}
        \toprule
        \textbf{Encoder} & \textbf{Description} \\
        \midrule
        Concat & $\mathbf{e} = \text{Linear}([\mathbf{e}^{(R)}, \mathbf{e}^{(S)}])$ \\
        Concat-MLP & $\mathbf{e} = \text{MLP}([\text{Linear}(\mathbf{e}^{(R)}), \mathbf{e}^{(S)}])$  \\
        Concat-MLP-Residual (w/o weights) & $\mathbf{e} = \text{MLP}([\text{Linear}(\mathbf{e}^{(R)}), \mathbf{e}^{(S)}]) + \mathbf{e}^{S}$  \\
        Concat-MLP-Residual & $\mathbf{e} = \text{MLP}([\text{Linear}(\mathbf{e}^{(R)}), \mathbf{e}^{(S)}]) + \alpha \mathbf{e}^{S}$ \\
        \bottomrule
    \end{tabular}
    }
\end{table}

\subsubsection{Decoder: ComplEx}
We choose ComplEx \citep{complex} as the decoder. For a triplet $(h, r, t)$, the encoder generates the embedding of $h$, $r$ and $t$, which are $\mathbf{e}_h \in \mathbb{R}$, $\mathbf{e}_r \in \mathbb{R}$, and $\mathbf{e}_t \in \mathbb{R}$, respectively. We then transform those embeddings from $\mathbb{R}^d$ to $\mathbb{C}^{d/2}$ by regarding the first $d/2$ dimensions as  the real part and the rest as the imaginary part.

Then, the score $f(h, r, t)$ of $(h, r, t)$ is computed by
\begin{align}
    f(h, r, t) = \text{Re}<\mathbf{e}_h, \mathbf{e}_r, \mathbf{e}_t>.
\end{align}

\subsection{Rule Miner}\label{sec:rule_miner}

Knowledge graph contains a wealth of structural information, and we can mine \emph{rules} from the knowledge graph. For example, we can mine the rule 
\begin{align*}
    \text{livesIn}(h, p) \land \text{marriedTo}(h,w) \Rightarrow \text{livesIn}(w,p)
\end{align*}

This rule captures the fact that the spouse of a person usually lives in the same place as the person \citep{amie+}. We can acquire such rules from the training data, and use them to enhance the performance of the inference model. 

We use the code of \href{https://github.com/lajus/amie}{AIME 3} to generate rules from the knowledge graph constructed by the training data. As the whole knowledge graph is too large, we sample five subgraphs from the whole graph. Then we apply the \href{https://github.com/lajus/amie}{AMIE 3} to generate rules from the five subgraphs, respectively. Finally, we merge all the rules to get the final rules.


        

After getting the generated rules, we use them to make prediction for unseen data. Suppose that the set of entities is $\mathcal{E} = \{ e_1, e_2, \ldots, e_{|\mathcal{E}|} \}$ and the set of relations is $\mathcal{R} = \{ r_1, r_2, \ldots, r_{|\mathcal{R}|} \}$. 
We define the adjacent matrix of $k$-th relation $r_k$ as $\textbf{M}_{r_k} \in \{0,1\}^{|\mathcal{E}\times \mathcal{E}|}$, where $[\textbf{M}_{r_k}]_{ij} = 1$ if and only if $(e_i, r_k, e_j)$ is a triple in the knowledge graph. 
For a rule $r_c (x,y) \Leftarrow r_a(x,y) \land r_b(y,z)$, the we can calculate the adjacent matrix of new triples for relation $r_c$ as follows. 
\begin{align*}
    \textbf{M}^N_{r_c} = \textbf{M}_{r_a} \textbf{M}_{r_b} - \textbf{M}_{r_c}
\end{align*}

The matrix $\textbf{M}^N_{r_c}$ contain the predictions of some unseen triples. Therefore, we can use the generated rules to promote the prediction of new triples. The calculations of other rules can be induced similarly.

\subsection{Inference Model}\label{sec:inference_model}
In this part, we introduce the inference model. In Section \ref{sec:inference_model_overall_architecture}, we introduce the overall architecture. In Section \ref{sec:inference_process}, we introduce the inference process. 

\subsubsection{Overall Architecture}\label{sec:inference_model_overall_architecture}
\begin{figure*}[h]
  \centering
  \includegraphics[scale=0.35]{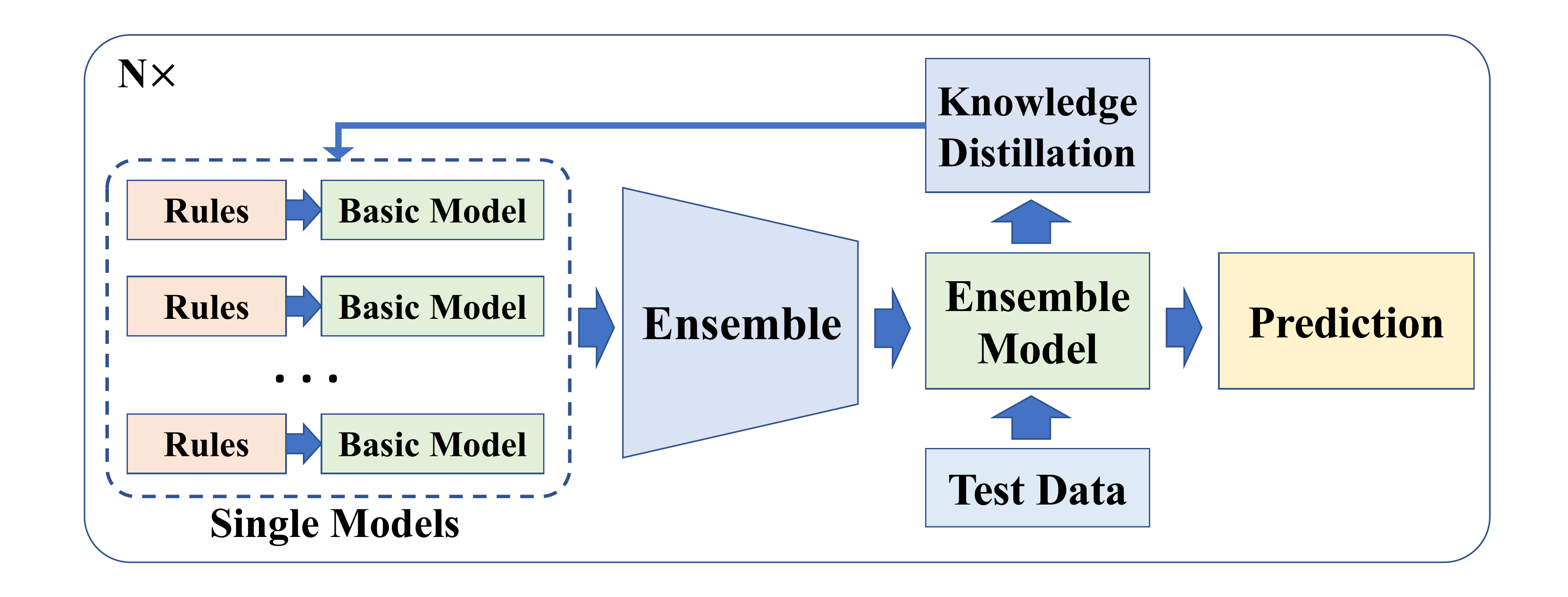}
  \caption{The overall architecture of our inference model.}
  \label{fig:inferece_overall_arch}
\end{figure*}

In this part, we introduce the design of our inference model, the overall architecture is shown in Figure \ref{fig:inferece_overall_arch}. The inference model takes the trained basic models, the mined rules and the test data as input, and it outputs the predictions on the test dataset. To make accurate predictions on the test dataset, the inference applies rule-based data augmentation, ensemble methods and knowledge distillation techniques. In the model, the inference procedure repeats several times, and the final prediction on the test set is the prediction of the last iteration.

\subsubsection{The Inference Process}\label{sec:inference_process}
\textbf{Rule-based Data Augmentation}\hspace{1.5mm} We filter the rules in the KG by their confidence, and use high-confident rules to generate new unseen triples (see Section \ref{sec:rule_miner}). We then use the newly generated triples to finetune the basic model.

\textbf{Ensemble}\hspace{1.5mm} Given $N$ trained single models, we apply average bagging to obtain a better ensemble model. Let $S_i \in \mathcal{R}^{|\mathcal{D}_{test}| \times n_c}$ denote the prediction of the $i$-th single model on the test set, where $|\mathcal{D}_{test}|$ denote the number of test samples, and $n_c$ denotes the number of candidate entities of each sample. When making predictions with the ensemble model on test data, the predictions of the ensemble model is 
\begin{align}
    S = \frac{\sum_{i=1}^{N} S_i}{N},
\end{align}
where $N$ is the number of single models for ensembling.


\textbf{Knowledge Distillation}\hspace{1.5mm} Since the ensemble model significantly outperforms single models, we use knowledge distillation \citep{distillation} to distill the superior performance of the ensemble model into single models. Specifically, we perform knowledge distillation by allowing the single models to learn the output of the ensemble model. In this way, the newly distilled single models could achieve similar performances to the previous ensemble model. We then repeat ensembling and distillation for several times. We use the predictions of the ensemble model at the last iteration as the final predictions of test data. Note that the knowledge distillation uses test tail candidates, which are unavailable in real-world machine learning problems, as models need to rank true tails among all the entities in practice. To evaluate our method in a more practical setting, we report the MRR of our model without knowledge distillation in Section \ref{sec:exp_inference_model}.


\section{Experiments}
In Section \ref{sec:train_protocol}, we introduce the training protocols. In Section \ref{sec:ablation_model}, we conduct the ablation studies on model design. In Section \ref{sec:exp_amie}, we present the details of rule mining. In Section \ref{sec:exp_inference_model}, we show the performance of the inference model on the validation dataset.

\subsection{Training Protocols} \label{sec:train_protocol}
\subsubsection{The Inverse Relation Setting}
We adopt the ``inverse relation setting'' for training. That is, we define an inverse relation $r^{-1}$ for each relation $r$ in the KG. Then, for each triplet $(h, r, t)$ in the training dataset, we add a new triplet $(t, r^{-1}, h)$ to the dataset. During training, each relation $r$ and its inverse relation $r^{-1}$ share the same RoBERTa embedding, but their shallow embeddings are different. 

After adding inverse relations, we only keep the ``tail mode`` during training. That is, for each triplet $(h, r, t)$, we only require the model to predict the tail entities $t$ given the query $(h, r, ?)$. Since inverse relations are introduced, the head prediction task is also included through the triplet $(t, r^{-1}, h)$ built from the inverse relation $r^{-1}$. 

\begin{table}[ht]
    \caption{The ablation studies on model design.}
    \label{table:model_ablation}
    \centering
    \resizebox{0.7 \columnwidth}{!}{
    \begin{tabular}{l *{3}{c}}
        \toprule
        \textbf{Decoder} & \textbf{Encoder} & \textbf{InvRel}  & \textbf{MRR (5\% Validation)} \\
        \midrule
        DistMult & Concat & No & 0.856 \\
        ComplEx & Concat & No & 0.852 \\
        ComplEx & Concat & Yes & 0.887 \\
        ComplEx & Concat-MLP & Yes & 0.905 \\
        ComplEx & Concat-MLP-Residual & Yes & 0.926 \\
        \bottomrule
    \end{tabular}
    }
\end{table}

\begin{table}[ht]
    \caption{The sampled subgraphs and the number of rules.}
    \label{table:subgraph_rules}
    \centering
    \resizebox{0.9 \columnwidth}{!}
    {
    \begin{tabular}{c *{4}{c}}
        \toprule
        \textbf{Subgraph ID} & \textbf{Sampled Triples} & \textbf{Number of Samples} & \textbf{Number of generated rules}\\
        \midrule
        0 & $\text{train\_hrt}[0: 200000000]$ & 200,000,000 & 7179 \\
        1 & $\text{train\_hrt}[200000000: 400000000]$ & 200,000,000 & 4981 \\
        2 & $\text{train\_hrt}[400000000:], \text{train\_hrt}[:100000000]$ & 201,160,482 & 8026 \\
        3 & $\text{train\_hrt}[100000000: 300000000]$ & 200,000,000 & 3903 \\
        4 & $\text{train\_hrt}[300000000:]$ & 201,160,482 & 5999 \\
        \bottomrule
    \end{tabular}
    }
\end{table}

\subsubsection{Hyperparameters}
The hyperparameters in our method are as follows.
\begin{itemize}
    \item The embedding dimension: 300
    \item The intermediate dimension of MLP: 3000
    \item Learning rate for shallow embeddings: 1e-1
    \item Learning rate for MLP parameters: 1e-4
    \item The number of processes: 4
    \item Batch size: 800
    \item Negative sample size: 100
\end{itemize}

\subsection{Ablation Studies on Model Design}\label{sec:ablation_model}
In this part, we conduct ablation studies on the model design. The design choices of encoders are listed in Table \ref{table:encoder_type}. The results are shown in Table \ref{table:model_ablation}. Experiments show that \textit{Concat-MLP-Residual} outperforms other encoders on the validation dataset. In Table 2, InvRel denotes the ``inverse relation`` setting, which is described in Section \ref{sec:train_protocol}.

\subsection{The Generated Rules by AMIE 3} \label{sec:exp_amie}

We use \href{https://github.com/lajus/amie}{AMIE 3} to mine rules from the knowledge graph. For computational efficiency, we sample five subgraphs from the whole graph and only mine rules  of length no longer than 3. We show the sampled subgraphs and the number of rules in Table \ref{table:subgraph_rules}, where we use train\_hrt to represent the NumPy array of the training triples. After getting the rules from the five subgraphs, we merge all the rules and finally get 11716 rules. We filter the rules by their confidence. For confidence greater than 0.95, there are 2062 rules. For confidence greater than 0.99, there are 1464 rules. 

\subsection{The Performance of Inference Model}\label{sec:exp_inference_model}

In this part, we conduct experiments on the inference model. We repeat ensembling and distillation for three times. Table \ref{table:exp_inference_model} shows the performance of the single model and the ensemble model on validation data. We use the ensemble model of Stage 3 to get the final predictions of the test data.

\begin{table}[ht]
    \caption{The MRR of the single model and the ensemble model on validation data at different inference stage.}
    \label{table:exp_inference_model}
    \centering
    {
    \begin{tabular}{c *{4}{c}}
        \toprule
        Model &  Stage 0 & Stage 1 & Stage 2 & Stage 3\\
        \midrule
        Single Model    &  0.926 & 0.970 & 0.973 & 0.976 \\
        Ensemble Model  &  0.953 & 0.973 & 0.977 & \textbf{0.978} \\
        \bottomrule
    \end{tabular}
    }
\end{table}

As discussed in Section \ref{sec:inference_process}, the knowledge distillation process uses the test tail candidates. To evaluate our method in a more practical setting, we conduct ablation studies by excluding the knowledge distillation process, and the results are shown in Table \ref{table:no_kd}. Experiments show that our model achieves an MRR of 0.9533 on the validation dataset without knowledge distillation (i.e., this model does not have any access to the provided test/val tail candidates).

\begin{table}[!h]
    \caption{The ablation studies on Knowledge Distillation (KD). Note that ``Ensemble Model without KD'' does not have any access to the provided test/val tail candidates.}\label{table:no_kd}
    \centering
    {
    \begin{tabular}{c *{4}{c}}
        \toprule
        Model &  MRR (Full Validation) & MRR (Full Test) \\
        \midrule
        Ensemble Model with KD    &  0.9782 & 0.9707  \\
        Ensemble Model without KD &  0.9533 & -  \\
        \bottomrule
    \end{tabular}}
\end{table}

\section{Conclusion}

In this paper, we introduce our proposed method for WikiKG90M-LSC track of KDD Cup 2021. In our method, we integrate three components---the basic model ComplEx-CMRC, the rule miner AMIE 3 and the inference model. In the inference model, we apply knowledge distillation by using test tail candidates, which are unavailable in the practical KG completion scenario. To evaluate our model in a more practical setting, we conduct ablation studies on knowledge distillation, and report the results that does not use val/test tail candidates. Experiments on the link prediction task demonstrate the effectiveness of our proposed method.

\bibliographystyle{unsrtnat}
\bibliography{references}  






\end{document}